\documentclass[conference,onecolumn,11pt]{IEEEtran}
\IEEEoverridecommandlockouts
\usepackage[margin=1in]{geometry}

\usepackage{cite}
\usepackage{amsmath,amssymb,amsfonts}
\usepackage{algorithmic}
\usepackage{algorithm}
\usepackage{graphicx}
\usepackage{textcomp}
\usepackage{xcolor}
\usepackage{booktabs}
\usepackage{subcaption}
\usepackage[misc]{ifsym}
\usepackage{float}
\usepackage{stfloats}
\fnbelowfloat

\usepackage{amsthm}
\newtheorem{theorem}{Theorem}

\theoremstyle{definition}

\usepackage{hyperref}
\def\BibTeX{{\rm B\kern-.05em{\sc i\kern-.025em b}\kern-.08em
    T\kern-.1667em\lower.7ex\hbox{E}\kern-.125emX}}

\begin{document}

\title{UCB-min: Combining Bandits Under Variance Heterogeneity\\ \large{(Extended Version)}\thanks{This is the extended version of the short paper of the same title accepted at IEEE BigData 2026.}}

\author{%
\IEEEauthorblockN{Lilia Izri}
\IEEEauthorblockA{\textit{LISN, CNRS} \\ \textit{Universit\'e Paris-Saclay}\\ Orsay, France \\ izri@lisn.fr}
\and
\IEEEauthorblockN{Beno\^it Groz}
\IEEEauthorblockA{\textit{LISN, CNRS} \\ \textit{Universit\'e Paris-Saclay}\\ Orsay, France \\ groz@lisn.fr}
\and
\IEEEauthorblockN{Silviu Maniu}
\IEEEauthorblockA{\textit{LIG, CNRS} \\ \textit{Univ. Grenoble-Alpes, Grenoble INP}\\ Grenoble, France \\ silviu.maniu@univ-grenoble-alpes.fr}
}

\maketitle

\begin{abstract}
Multi-armed bandits are a natural framework for many real-world sequential decision problems, from recommender systems and clinical trials to educational test recommendation and online advertising. The performance of these algorithms depends on how they explore different arms, and exploration strategies range from variance-independent (e.g., UCB1) to variance-aware (e.g., UCB-V). In practice the reward variance of each arm is rarely known and varies widely across arms, and the reward support is often underspecified. In these cases no single rule is uniformly best: variance-independent methods are robust but loose when variances are small, while variance-aware methods are tight only when their variance estimates and the assumed support bound are reliable. Such heterogeneous, unknown variances arise in many settings, for instance when multiple objectives are scalarized into a single reward. We propose UCB-min, a per-arm adaptive meta-algorithm that selects, for each arm, the tighter of a variance-independent and a variance-aware exploration bonus, without prior knowledge of the variance regime and without reward normalization. We prove that, with UCB-V as variance-aware base, its regret is bounded by the sum of the two base bounds at the standard exploration constant, and by their per-arm minimum, up to a constant, for a larger one. We validate it on synthetic data, an educational tutoring task, and two group-recommendation benchmarks (MovieLens, Steam), showing that UCB-min consistently tracks the best base algorithm across diverse variance regimes and datasets.
\end{abstract}


\section{Introduction}\label{sec:introduction}
Multi-armed bandit (MAB) algorithms are fundamental to sequential decision-making under uncertainty, with applications in clinical trials~\cite{villar2015multi,aziz2021multi}, recommender systems~\cite{li2010contextual,da2022multi}, resource allocation~\cite{zuo2021combinatorial}, and online advertising~\cite{schwartz2017customer,bouneffouf2019survey}. UCB methods assign each arm an optimistic index combining its empirical mean with an exploration bonus and pull the arm with the highest index; these bonuses range from variance-independent (UCB1~\cite{auer2002finite}) to variance-aware (UCB-V~\cite{audibert2007tuning,fan2024precise}) design. The formal definitions and regret comparisons are given in Section~\ref{sec:problem}.

In practice the reward variance of each arm is rarely known and varies widely across arms, so neither family is uniformly best: variance-independent methods are robust but loose when variances are small, while variance-aware methods are tight only when their variance estimates and the assumed support bound are reliable. Normalization could remove this scale dependence but is often impractical online (see Section~\ref{sec:challenge}).

This occurs in many real-world applications, in particular inherently \textbf{multi-objective} ones: recommender systems trade off accuracy with diversity, novelty, and fairness~\cite{lacerda2017multi,wang2022fairness}, clinical trials optimize efficacy and safety~\cite{aziz2021multi}, and advertising platforms balance user satisfaction and diversity~\cite{mehrotra2020multi}. In the recommendation settings we consider, each arm is an item (e.g., a news article or product), pulling it recommends that item, and the reward vector's dimensions capture properties such as relevance, clicks, ratings, or engagement. A common way to handle several objectives is \textbf{scalarization}: combining the reward dimensions into a single scalar via application-specific weights $w \in \mathbb{R}^D$, after which standard single-objective bandits such as UCB apply. This reduction is itself a source of the variance heterogeneity above, since the scalarized variance depends on the weights and the cross-objective covariance, quantities often unknown in advance, which makes scalarized multi-objective bandits a natural motivation for our work.

To address this, we propose \textbf{UCB-min}, a per-arm adaptive meta-algorithm for scalarized multi-armed bandits that selects, for each arm, the tighter of a variance-independent and a variance-aware exploration bonus. This choice requires no prior knowledge of the variance regime and no reward normalization. The approach is validated on synthetic datasets and real-world multi-objective applications: an \textbf{education} dataset~\cite{bouarour2024multi}, where multi-objective formulations and algorithm choice matter depending on the variance regime, and \textbf{group recommendation} datasets~\cite{amer2009group}, where per-arm adaptation provides robustness across arms with different variance profiles.

\paragraph*{Contributions}
Our main contributions are the following:
\begin{enumerate}
  \item We analyze the effect of variance scales on regret in multi-objective bandits and empirically show that the relative performance of variance-independent (UCB1) and variance-aware (UCB-V) algorithms is data dependent.
  \item We propose per-arm adaptive meta-algorithms that dynamically select between variance-independent and variance-aware exploration bonuses on a per-arm basis, without prior knowledge of the variance regime and without normalization.
  \item We provide a theoretical guarantee for the per-arm minimum rule with UCB-V (UCB-minV1): the cumulative expected regret has the same asymptotic bound as the original algorithms and, for a larger exploration constant, tracks the regret of the best algorithm (Theorem~\ref{thm:main}); the variant with UCB-V2 (UCB-minV2) is validated empirically.
  \item We validate our approach on synthetic, educational, and recommendation datasets, showing robust empirical improvements across diverse variance regimes.
\end{enumerate}

\section{Multi-Objective Multi-Armed Bandits}\label{sec:problem}

This section introduces the (multi-objective) bandit framework and the UCB
algorithms used throughout the paper. We first define the general stochastic
bandit setting with multiple objectives, then present linear scalarization as a
means of reducing it to a single-objective problem, and discuss the challenges
this reduction creates for algorithm selection.

\subsection{Multi-Objective Bandit Framework}

We consider a stochastic multi-armed bandit with $K$ arms
$\mathcal{A}=\{1,\dots,K\}$ and $D$ objectives. At each round $t=1,\dots,T$, the bandit algorithm selects an arm $a_t\in\mathcal{A}$
and observes a reward vector $\mathbf{r}_t\in\mathbb{R}^D$. For each arm $a$, the
rewards are i.i.d.\ with unknown mean vector $\boldsymbol{\mu}_a\in\mathbb{R}^D$ and
covariance matrix $\mathbf{\Sigma}_a\in\mathbb{R}^{D\times D}$. Let $N_a(t)=\sum_{\tau=1}^t\mathbf{1}\{a_\tau=a\}$ be the number of times arm
$a$ has been pulled up to time $t$. The empirical estimators are, with the variance normalized by $N_a(t)$ as in~\cite[Theorem~1]{audibert2009exploration}:
\begin{align*}
\hat{\boldsymbol{\mu}}_a(t)&=\frac{1}{N_a(t)}\sum_{\tau=1, a_\tau=a}^t\mathbf{r}_\tau,\\
\hat{\mathbf\Sigma}_a(t)&=\frac{1}{N_a(t)}\sum_{\tau=1, a_\tau=a}^t
\bigl(\mathbf{r}_\tau-\hat{\boldsymbol{\mu}}_a(t)\bigr)
\bigl(\mathbf{r}_\tau-\hat{\boldsymbol{\mu}}_a(t)\bigr)^\top.
\end{align*}

\paragraph*{Pareto vs.\ scalarized}
A natural performance criterion in multi-objective bandits is Pareto regret, which measures suboptimality with respect to the Pareto front, which is the set of arms not simultaneously dominated by any other in all objectives~\cite{drugan2013designing,turgay2018multi}.
However, identifying and tracking the Pareto front can be computationally expensive. We therefore focus on \emph{scalarized regret}: given application-specific weights, the objectives are combined into a single scalar, reducing the problem to a standard single-objective bandit that is directly compatible with existing UCB-style algorithms and well-suited when the desired trade-off is known in advance.

Linear scalarization reduces the multi-objective problem to a standard
single-objective stochastic bandit, enabling the direct use of well-studied UCB
algorithms and their theoretical guarantees. Given a fixed weight vector
$\mathbf{w}\in\mathbb{R}^D$ with $\mathbf{w}_d\geq 0$ and $\sum_{d=1}^D \mathbf{w}_d=1$, the scalarized reward observed at round $t$ is $X_t = \mathbf{w}^\top \mathbf{r}_t$. For each
arm~$a$, the scalarized mean and variance are:
\begin{equation}\label{eq:scalar_var}
\mu_a^{(s)} = \mathbf{w}^\top\boldsymbol{\mu}_a,\qquad
\sigma_{a}^{2,(s)} = \mathbf{w}^\top\Sigma_a \mathbf{w},
\end{equation}

where the superscript $(s)$ marks scalarized quantities.

After $t$ rounds, the empirical estimators are:

\begin{equation}\label{eq:scalarized_params}
\hat\mu_a^{(s)}(t) = \mathbf{w}^\top\hat{\boldsymbol{\mu}}_a(t),\qquad
\hat\sigma_a^{2,(s)}(t) = \mathbf{w}^\top\hat{\Sigma}_a(t) \mathbf{w}.
\end{equation}

\subsection{UCB Algorithms and Regret}
\label{sec:ucb-algos}

The goal of bandit algorithms is to minimize the cumulative regret. Let
$a^\star=\arg\max_a \mu_a^{(s)}$,
$\mu_\star^{(s)}=\mu_{a^\star}^{(s)}$, and
$\Delta_a=\mu_\star^{(s)}-\mu_a^{(s)}$. The expected cumulative regret is:
\begin{equation}\label{eq:regret}
R(T) = \mathbb{E}\Bigl[\sum_{t=1}^T
\bigl(\mu_\star^{(s)}-X_{t}\bigr)\Bigr]
= \sum_{a\in\mathcal{A}} \Delta_a \mathbb{E}[N_a(T)].
\end{equation}

To minimize this quantity, UCB-type algorithms assign each arm an index combining the empirical scalarized mean with an exploration bonus and select the arm with the highest index. $\beta^{(\cdot)}_a(t)$ denotes a pure \emph{exploration bonus} and $B^{(\cdot)}_a(t) = \hat\mu_a^{(s)}(t) + \beta^{(\cdot)}_a(t)$ the corresponding \emph{full index}; the algorithm pulls $\arg\max_a B^{(\cdot)}_a(t)$.
This work builds upon well-known algorithms (presented below) which we apply to the scalarized multi-objective setting for rewards having support $[0,b]$. 

\paragraph*{UCB1~\cite{auer2002finite}}
UCB1 is variance-independent; its bonus and index are
\begin{equation}\label{eq:ucb1}
\beta^{(1)}_a(t) = b\sqrt{\frac{2\log t}{N_a(t)}},
\quad
B^{(1)}_a(t) = \hat\mu_a^{(s)}(t) + \beta^{(1)}_a(t),
\end{equation}
achieving regret
\[
\sum_{a:\Delta_a>0} \frac{8b^2\log T}{\Delta_a}
+ \Bigl(1+\frac{\pi^2}{3}\Bigr)\sum_a \Delta_a.
\]
When $b=1$ this reduces to the standard formulation. Two refinements
from the same work are UCB1-Tuned and UCB1-Normal~\cite{auer2002finite}.
UCB1-Tuned replaces the exploration bonus by
\[
\sqrt{\frac{\log t}{N_a(t)}\min \left\{\frac{1}{4},\, \hat\sigma_a^{2,(s)}(t)+\sqrt{\frac{2\log t}{N_a(t)}}\right\}},
\]
using an empirical variance estimate to tighten the bonus; it often
improves empirical performance but lacks formal regret guarantees. 
UCB1-Normal~\cite{auer2002finite} targets Gaussian rewards; in our case, this assumption may not hold after scalarization of objectives.

\paragraph*{UCB-V~\cite{audibert2007tuning}}
UCB-V explicitly uses variance in the exploration bonus:
\begin{equation}\label{eq:ucbv}
\beta^{(V)}_a(t) = \sqrt{\frac{2\hat\sigma_a^{2,(s)}(t) c\log t}{N_a(t)}}
  + \frac{3cb\log t}{N_a(t)},
\end{equation}
with $B^{(V)}_a(t) = \hat\mu_a^{(s)}(t) + \beta^{(V)}_a(t)$ and $c=1.2$. Its regret bound~\cite[Theorem~4]{audibert2009exploration} is
\[
10\sum_{a:\Delta_a>0} \left(\frac{\sigma_a^{2,(s)}}{\Delta_a} + 2b\right)\log T,
\]
which improves over UCB1 when per-arm variances are small relative to~$b$.

\paragraph*{UCB-V2~\cite{fan2024precise}}
UCB-V2, our name for the version of UCB-V analyzed in~\cite{fan2024precise}, refines UCB-V with the bonus
\begin{equation}\label{eq:ucbv2}
\beta^{(V2)}_a(t) =\max \left(
  \frac{\hat\sigma_a^{(s)}(t)}{\sqrt{\rho\log T}}, 
  \frac{b}{\sqrt{N_a(t)}}
\right)\cdot\frac{\rho\log T}{\sqrt{N_a(t)}},
\end{equation}
and $B^{(V2)}_a(t) = \hat\mu_a^{(s)}(t) + \beta^{(V2)}_a(t)$, where $\rho$ is a tuning parameter ($\rho=2$ in our experiments). When $b=1$, $T\geq 3$, and $\rho\geq C_0$ for a universal constant $C_0$, with probability at least $1-3K/T^2$, its regret scales as
\[
16\min \left\{
  \sqrt{\sum_{a=2}^K \sigma_a^{2,(s)}},\,
  \sum_{a=2}^K \frac{16\sigma_a^{2,(s)}}{\sigma_1^{(s)}}
\right\}\sqrt{\rho T\log T}
+ (K-1)\sqrt{\rho\log T},
\]
where the arms are indexed so that the optimal arm $a^\star = 1$.
\section{Other Related Work}\label{sec:related-work}

We review related work on MAB alternatives and on
meta-combination strategies for bandit algorithms beyond the UCB-based methods already introduced in Section~\ref{sec:ucb-algos}.

\paragraph*{MAB alternatives}
Thompson Sampling (TS)~\cite{thompson1933likelihood,agrawal2012analysis} and
KL-UCB~\cite{garivier2011kl} often perform well but rely on distributional assumptions: TS needs a prior and likelihood model, and KL-UCB derives bounds from KL-divergences for specific families (e.g., Bernoulli); EUCBV~\cite{mukherjee2018efficient} combines variance estimates with arm elimination. After scalarization, which mixes heterogeneous objectives (e.g., binary and continuous), the reward distribution is unknown in advance, so we focus on distribution-free UCB methods and do not compare against TS or KL-UCB, but against UCB1, UCB1-Tuned, UCB-V, UCB-V2, and EUCBV (Section~\ref{sec:experiments}).

\paragraph*{Multi-objective problems}
Multi-objective sequential decision problems are addressed either through Pareto-based methods, which maintain the set of non-dominated solutions, or through scalarization-based methods, which reduce the objectives to a single scalar~\cite{drugan2013designing,roijers2013survey}. Pareto-based approaches require repeated dominance checks, and the Pareto front can be prohibitively large~\cite[Sec.~4.2.2]{roijers2013survey}, which has motivated cost-reducing heuristics such as annealing~\cite{yahyaa2014annealing}. Scalarization (linear, Chebyshev, randomized, etc.) is a lightweight alternative, directly compatible with UCB/Thompson-style bandit and RL algorithms. Studies include a MOMAB framework with empirical comparisons~\cite{drugan2013designing}, contextual Pareto regret bounds~\cite{turgay2018multi}, the Generalized Gini Index~\cite{busa2017multi}, and, in multi-objective RL, a survey of MOMDP/MORL methods with conditions under which scalarization preserves optimality~\cite{roijers2013survey} and the learning of policies that treat several users equitably~\cite{siddique2020learning}. As discussed in Section~\ref{sec:introduction}, scalarization induces heterogeneous per-arm variance, which motivates the per-arm rule validated in Section~\ref{sec:experiments}.

\paragraph*{Meta algorithms}
Several works consider meta-combination of bandit strategies. 
Meta-UCB~\cite{cutkosky2020upper} treats each base learner as a single arm in a
higher-level bandit, and its regret depends only on that of the best base learner. Corralling~\cite{arora2021corralling} adaptively
selects which base algorithm to run each round.
\cite{pmlr-v235-cheung24a} proposes MIN-UCB, a per-arm min-of-indices
approach combining online samples with offline data via a bias-correction
term, providing instance-dependent and near-minimax guarantees but requiring offline data
and a known bias bound. None of these works studies a fully online, distribution-free per-arm rule selecting between variance-independent and variance-aware indices without offline data or known bias bound, a gap this work addresses. These are therefore not directly comparable baselines: Meta-UCB and Corralling select among \emph{entire} algorithms rather than per arm, MIN-UCB additionally requires offline data and a known bias bound, and adversarial selectors such as Exp3 ignore variance entirely. A related idea appears in the stopping-rule literature, where \cite{mnih2008efficient} combines a variance-independent and a variance-aware stopping rule by stopping at the earlier of the two; that work targets $(\epsilon,\delta)$ mean estimation rather than per-arm bandit regret minimization.
Another family of methods scales bandits to large recommendation settings by sharing statistics across users: CLUB~\cite{gentile2014club} and its data-dependent refinement~\cite{li2015datadependent} cluster users online, DCCB~\cite{korda2016distributed} distributes the clustering over a peer-to-peer network, and DistCLUB~\cite{distclub2020} constructs clusters
lazily to reduce communication. These are contextual linear bandits requiring item features and the identity of the user at each round, hence not comparable baselines in the featureless scalarized setting we consider: they exploit similarity across users, whereas UCB-min adapts to per-arm variance.

\section{UCB-min: Per-Arm Adaptive Meta-Algorithm}\label{sec:meta}

\subsection{Variance Changes in Multi-Objective Scalarization}\label{sec:challenge}

Eq.~\eqref{eq:scalar_var} shows that the scalarized variance depends on the weights and on the covariance structure of the objectives, and correlation effects can reduce it substantially. Scalarized variances may therefore range from very small to comparable to the largest single-objective variance, creating a dilemma between variance-aware methods (effective when variance is small) and variance-independent ones (more robust otherwise).

Two practical difficulties make this dilemma worse. First, \emph{normalization} is often impractical: reward scales are unknown online, non-stationary settings require continuous re-estimation, and for physical objectives (energy, time, cost) no predetermined upper bound exists. It does not reliably reduce regret either (see Figs.~\ref{fig:norm_raw}-\ref{fig:ml_normalized}). Second, the algorithms above require the support bound $b$, which is generally unknown without normalization: Gaussian rewards are unbounded and, for real-world data, the effective range is hard to determine in advance. Using a large $b$ inflates all exploration bonuses and leads to near-linear regret (see Fig.~\ref{fig:b5}).

\subsection{The UCB-min Meta-Algorithm}\label{sec:ubc-min}

To control these issues, we propose \textsc{UCB-min}, an adaptive per-arm \emph{meta}-algorithm that selects, for each arm, the tighter of a variance-independent and a variance-aware exploration bonus. It is analyzed in Theorem~\ref{thm:main}, validated in Section~\ref{sec:experiments}, and, although both bonuses still, in theory, depend on the support bound $b$, is empirically robust to an underestimated $b$, allowing the default $b=1$ across variance regimes. Throughout, ``meta'' refers to combining existing base algorithms in this sense, and not to meta-learning.

\begin{algorithm}[h]
\caption{UCB-min}
\label{alg:meta}
\begin{algorithmic}[1]
\REQUIRE Arms $\mathcal{A}$, horizon $T$, weight vector $\mathbf{w}$
\STATE Initialize: $\forall a$: $N_a \gets 0$,
  $\hat{\boldsymbol{\mu}}_{a} \gets \mathbf 0_D$, $\hat{\mathbf{\Sigma}}_{a} \gets  \mathbf 0_{D\times D}$
\STATE Pull each arm $a\in\mathcal A$ once.
\FOR{$t = 1$ to $T$}
    \FOR{each $a \in \mathcal{A}$}
        \STATE Compute $\hat\mu_a^{(s)}(t)$ and
          $\hat\sigma_a^{2,(s)}(t)$ (Eq.~\eqref{eq:scalarized_params})
        \STATE Compute the bonuses $\beta^{(1)}_a(t)$ (Eq.~\eqref{eq:ucb1}) and
          $\beta^{(\mathrm{var})}_a(t)$ (Eq.~\eqref{eq:ucbv} or \eqref{eq:ucbv2})
        \STATE $B_a(t) \gets \hat\mu_a^{(s)}(t)
          + \min\{\beta^{(1)}_a(t),  \beta^{(\mathrm{var})}_a(t)\}$
    \ENDFOR
    \STATE Pull $a_t \gets \arg\max_a\, B_a(t)$, observe
      $\mathbf{r}_{t}$, update $N_{a_t}$, $\hat{\boldsymbol{\mu}}_{a_t}$, $\hat{\mathbf{\Sigma}}_{a_t}$
\ENDFOR
\end{algorithmic}
\end{algorithm}

For each arm~$a$, we compute both the variance-independent bonus
$\beta^{(1)}_a(t)$ (Eq.~\eqref{eq:ucb1}) and a variance-aware bonus $\beta^{(\mathrm{var})}_a(t)$
(Eq.~\eqref{eq:ucbv} or~\eqref{eq:ucbv2}), keep the tighter one, and add the empirical mean once:
\begin{equation}\label{eq:per-arm-min}
B_a(t) = \hat\mu_a^{(s)}(t) + \min \big\{\beta^{(1)}_a(t),\, \beta^{(\mathrm{var})}_a(t)\big\}
       = \min \big\{B^{(1)}_a(t),\, B^{(\mathrm{var})}_a(t)\big\}.
\end{equation}
Both forms coincide since $\hat\mu_a^{(s)}(t)$ is shared, and Appendix~\ref{apx:proof} uses the second one, which reasons directly on indices. Each arm thus benefits from variance-aware exploration when its empirical variance is low, and reverts to UCB1 when the variance-aware bonus would be too large. Both bonuses are already valid upper confidence bounds, so their minimum is valid as well, and being the tighter of the two it avoids the exploration that a larger combination (maximum, mean) would incur for no gain in guarantee. This is especially useful when~$b$ is misspecified: for a small $b$ (e.g., the default $b=1$), $\beta^{(1)}_a(t)$ remains tight while the variance-aware bonus may still be inflated by large empirical variances. The selection also depends on the pull count $N_a(t)$: the variance-aware bonus carries an extra $O(1/N_a(t))$ term that dominates in early iterations, so the minimum then falls back on $\beta^{(1)}_a(t)$, and the variance plays its role mainly once $N_a(t)$ is large enough.

We consider two variants depending on the variance-aware base:
\begin{itemize}
  \item \textbf{UCB-minV1}: $\beta^{(\mathrm{var})}_a = \beta^{(V)}_a$, Eq.~\eqref{eq:ucbv},
  \item \textbf{UCB-minV2}: $\beta^{(\mathrm{var})}_a = \beta^{(V2)}_a$, Eq.~\eqref{eq:ucbv2}.
\end{itemize}
Both share the same structure, presented in
Algorithm~\ref{alg:meta}. At each round, we compute scalarized
statistics
$\hat\mu_a^{(s)}(t) = \mathbf{w}^\top\hat\mu_a(t)$ and
$\hat\sigma_a^{2,(s)}(t) = \mathbf{w}^\top\mathbf{\hat\Sigma}_a(t) \mathbf{w}$, evaluate both
bonuses, and pull
$\arg\max_a B_a(t)$ with $B_a(t)$ as in Eq.~\eqref{eq:per-arm-min}.

\paragraph*{Computational cost}
UCB-min evaluates two bonuses per arm at each round, so its cost per round is of the same order as that of UCB1 and UCB-V, namely $O(K)$ given the scalarized statistics. With a fixed weight vector $\mathbf{w}$, these statistics are the empirical mean and variance of the scalarized rewards observed on each arm, since
\[
\mathbf{w}^\top\hat{\mathbf\Sigma}_a(t)\mathbf{w}=\frac{1}{N_a(t)}\sum_{\tau\leq t,\, a_\tau=a}\bigl(X_\tau-\hat\mu_a^{(s)}(t)\bigr)^2,
\]
and can be updated online in $O(1)$ time and memory per arm; maintaining the full covariance $\hat{\mathbf\Sigma}_a(t)$, at a cost of $O(D^2)$ per arm, is only needed to evaluate several weight vectors on the same observations, where diagonal or low-rank approximations would reduce this cost for many objectives.

\paragraph*{Regret upper bound}

We next show that the asymptotic regret bound of UCB-minV1 is the same as the regret bound of the original algorithms:

\begin{theorem}

\label{thm:main}

Fix per-arm $\Delta_a$, $\sigma_a^2$, and support bound $b$. The statement holds for any bounded rewards, so we omit the $^{(s)}$ superscript here and in the proof.
\begin{enumerate}
\item[(i)] For $c\geq 4$ in the UCB-V index, UCB-minV1 attains for each arm the \emph{minimum} of the UCB1 and UCB-V regret bounds, up to a constant:
\[O\left(\left(\sum_a\min\bigl(\tfrac{\sigma_a^2}{\Delta_a} + b,\, b^2/\Delta_a\bigr) \right)\log T +\sum_a \Delta_a\right),\]
i.e.\ it is never worse than the better of the two on any arm.
\item[(ii)] At the standard $c=1.2$~\cite{audibert2007tuning}, UCB-minV1 still enjoys the weaker guarantee of being bounded by the \emph{sum} of the two base bounds:
\[O\left(\left(\sum_a\frac{\sigma_a^2+b^2}{\Delta_a} + b \right)\log T +\sum_a \Delta_a\right).\]
\end{enumerate}
\end{theorem}
 We use $c=1.2$ in our experiments, the value fixed by~\cite{audibert2007tuning}, at which only part~(ii) is proved; a larger $c$ would keep the guarantee valid but enlarges the bonus, hence adds exploration, which is not always desirable (e.g. Fig.~\ref{fig:b5}).  While our theoretical guarantees are limited to UCB-minV1 and the stronger guarantee of part (i) also requires a larger $c$, our experiments in Section~\ref{sec:experiments} provide empirical evidence for similar behavior beyond the scope of the theorem, with both UCB-minV1 and UCB-minV2 appearing to track the best of their base algorithms.

\paragraph*{Note}\label{note:global}

An alternative is \emph{global} selection: at each round, estimate the regret upper bounds of both UCB1 and the variance-aware algorithm, then apply the one with the smaller bound to \emph{all} arms (\textbf{Meta-1} with UCB-V, \textbf{Meta-2} with UCB-V2). However, it ignores heterogeneous per-arm variance, accumulates early-round estimation errors, and relies on bounds that may greatly overestimate actual regret; our experiments confirm it is unreliable, so we focus on the per-arm rule.

\section{Experiments}\label{sec:experiments}

We present the experimental results, with two objectives: (1) show how \textsc{UCB-min} adapts to different variance regimes, and (2) show that real-world multi-objective applications benefit from this adaptation.

\subsection{Datasets}
We evaluate on three types of data: synthetic instances, an educational tutoring simulator, and two group recommendation benchmarks (MovieLens, Steam).

\textbf{\textit{Synthetic Data.}} We generate instances with $K = 20$ arms and $D = 2$ independent Gaussian objectives, with means in $[0,1]$ and standard deviations in $[\sigma_{\min}, \sigma_{\max}]$. Two dimensions suffice, as the goal is to validate behavior under controlled variance regimes.

\href{https://eedischool.com/projects/neurips-education-challenge}{\textbf{\textit{Educational Data.}\footnote{\url{https://eedischool.com/projects/neurips-education-challenge}}}}
The dataset~\cite{wang2021results} contains 4,918 students, 948 questions, and 1.38M responses. Each question is assigned a difficulty level $l \in \{1,\dots,5\}$ based on global success rate; each student's proficiency $l_u$ is the highest level with $\geq 50\%$ correct answers. Arms correspond to the $K=5$ difficulty levels. Following~\cite{bouarour2024multi}, pulling arm $a$ (level $l_a$) for a student at level $l_u$ yields the following reward components that all lie in $[0, 1]$:
\begin{itemize}
    \item \textbf{Aptitude (A)}: $r^{(A)}$ a binary reward indicating correct/incorrect answer.
    \item \textbf{Competence (C)}: $r^{(C)} = 1 - |l_u - l_a|/K$ (difficulty matching).
    \item \textbf{Performance (P)}: $r^{(P)} = \max\{0, (l_a - l_u)/K\}$ (potential progression).
    \item \textbf{Gap (G)}: $r^{(G)} = 1 - |l_{\text{fail}} - l_a|/K$ (recovery from failure with $l_{\text{fail}}$ the user's failing level).
\end{itemize}

\textbf{\textit{\href{https://kaggle.com/datasets/antonkozyriev/game-recommendations-on-steam}{Steam}\footnote{\url{https://kaggle.com/datasets/antonkozyriev/game-recommendations-on-steam}} and \href{https://grouplens.org/datasets/movielens/}{MovieLens 25M}\footnote{\url{https://grouplens.org/datasets/movielens/}}: Group Recommendations.}}
MovieLens 25M~\cite{harper2015movielens} contains 25M ratings (scale 0.5-5) from 162K users on 62K movies across 20 genres. Steam contains 41M recommendations from 14M users on 51K games with 337 tag categories; we use an extract of 4M user profiles. MovieLens groups of 20 users are formed via a variance-constrained process, whereas Steam groups are real community groups extracted via the Steam API. The relevance of a category for a group is the average rating over group members and items in that category. Objectives follow~\cite{amer2009group}; arms are categories and pulling category $c$ for group $G$ yields:
\begin{itemize}
    \item \textbf{Relevance}: $r^{(R)}_{c,t} = \mathrm{rel}(u, c)$ for randomly sampled user $u \in G$.
    \item \textbf{Pairwise Disagreement}: $r^{(D)}_{c,t} = |\mathrm{rel}(u,c) - \mathrm{rel}(v,c)|$ for sampled pair $u, v$.
\end{itemize}
Disagreement is inverted to $r_{\max} - r^{(D)}$ for maximization, where $r_{\max}$ is the maximum rating of the dataset. Rewards lie in $[0,5]$ for MovieLens (not normalized) and $[0,1]$ for Steam.

\subsection{Evaluation Metrics}

The algorithms are evaluated along three categories of metrics, formally defined in Appendix~\ref{apx:metrics}.

\textbf{Regret:} cumulative regret, as defined in Eq.~\eqref{eq:regret}.

\textbf{Ranking quality:} Hit Rate@$k$, assessing whether one of the last $k$ recommended arms is among the 5 arms with the highest true scalarized mean.

\textbf{Domain-specific:} Click-Through Rate (CTR) for MovieLens and Steam; skill progression (NCC) for the educational task.

\subsection{Implementation and Experiments Details}

\paragraph*{Machine specifications}
All experiments ran on a Linux server (8-core Intel Xeon E5-2609v2 2.5GHz, 64GB RAM).

\paragraph*{Experiments} Each algorithm is run 50 times and averaged, under identical seeds, horizon and value of $b$; all objectives get equal weights. We aggregate over \textbf{200~users} (diverse skill levels) for Education and \textbf{50~groups} for Steam and MovieLens, the latter formed to ensure variance heterogeneity. We report the algorithms of Sections~\ref{sec:problem}-\ref{sec:meta}; on the real-world datasets, EUCBV and Meta-1 are omitted for clarity as they never reach the lowest regret on synthetic data, and UCB-minV1 as the experiments focus on UCB-minV2; their full curves are available online.\textsuperscript{\ref{fn:github}} Error bars report one standard deviation over the runs; they are omitted in some figures for readability. The open-source Python implementation, datasets descriptions and supplementary materials are available online.\href{https://github.com/LiliZr/MoMAB}{\footnote{\label{fn:github}\url{https://github.com/LiliZr/MoMAB}}} The raw experimental results, additional results and datasets are also available online.\href{https://bit.ly/4s2nANs}{\footnote{\url{https://bit.ly/4s2nANs}}}

\paragraph*{Support bound}
We set $b=1$ everywhere. This matches the true support for the educational and Steam data (rewards in $[0,1]$); for MovieLens and high-variance synthetic data it underestimates the support, matching our setting where the reward range is unknown in advance. The theoretical guarantees of all compared methods no longer apply there; this is exactly the purpose of this evaluation, which assesses their empirical robustness to an underestimated $b$. Fig.~\ref{fig:b5} reports a well-specified bound for MovieLens.

\subsection{Results}

\paragraph*{Variance regimes - Synthetic data}

\begin{figure}[H]
  \centering
  \setlength{\abovecaptionskip}{3pt}
  \setlength{\belowcaptionskip}{0pt}
  \includegraphics[width=\textwidth]{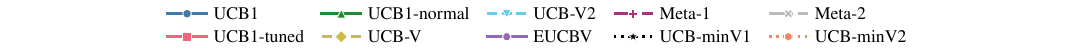}\\[2pt]
  \begin{subfigure}[t]{0.325\textwidth}
    \centering
    \includegraphics[width=\linewidth]{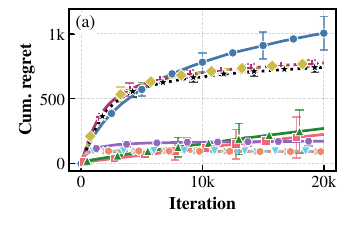}%
    \phantomsubcaption\label{fig:synth_small}
  \end{subfigure}\hfill
  \begin{subfigure}[t]{0.325\textwidth}
    \centering
    \includegraphics[width=\linewidth]{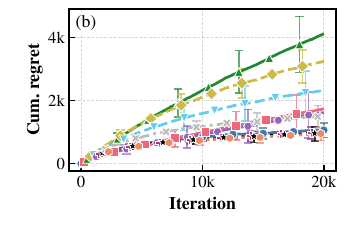}%
    \phantomsubcaption\label{fig:synth_big}
  \end{subfigure}\hfill
  \begin{subfigure}[t]{0.325\textwidth}
    \centering
    \includegraphics[width=\linewidth]{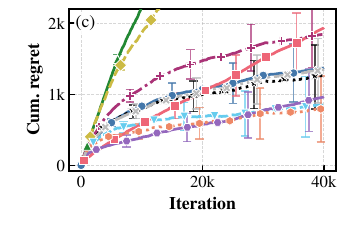}%
    \phantomsubcaption\label{fig:synth_arms_regret}
  \end{subfigure}
  \caption{Cumulative regret across variance regimes: (a) $\sigma \in [0.1, 0.3]$, (b) $\sigma \in [2, 4]$, (c) heterogeneous arms. (a) and (b) share the same scalarization weights, whereas (c) illustrates the impact of heterogeneous arm variances (high variance for half of the arms, low variance for the other half); its y-axis is cut at 2.2k (UCB-V and UCB1-Normal end at 3.7k and 6.5k).}
  \label{fig:synthetic}
\end{figure}

\begin{figure}[H]
  \centering
  \setlength{\abovecaptionskip}{3pt}
  \includegraphics[width=0.6\textwidth]{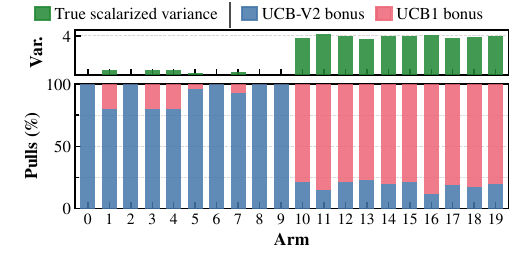}
  \caption{Per-arm selection on the heterogeneous instance of Fig.~\ref{fig:synth_arms_regret}. Top: true scalarized variance per arm. Bottom: how the pulls of each arm split between the two bonuses.}
  \label{fig:synth_arms_selection}
\end{figure}

Figures~\ref{fig:synth_small}-\ref{fig:synth_big} show cumulative regret under two variance
regimes with the same scalarization weights. In the low-variance regime
(Fig.~\ref{fig:synth_small}), UCB-V2 achieves roughly $10\times$ lower
final regret than UCB1, confirming the theoretical advantage of
variance-aware bounds when $\sigma_a^{2,(s)}$ is small. In the
high-variance regime (Fig.~\ref{fig:synth_big}), this relationship
reverses: UCB-V2 shows near-linear regret growth, as large empirical
variances make its bonus too large and cause over-exploration. Since Gaussian rewards exceed $[0,1]$ in this regime,  $b=1$ underestimates
the true support; nonetheless, UCB1's bonus is smaller than UCB-V2's variance-dependent bonus, making UCB1 more robust
in this regime. UCB-minV1 adapts automatically, tracking UCB-V in Fig.~\ref{fig:synth_small} and UCB1 in Fig.~\ref{fig:synth_big} without prior knowledge; UCB-minV2 behaves similarly but with lower regret thanks to the tighter UCB-V2 bound, so the remaining experiments focus on it, although Theorem~\ref{thm:main} does not cover it. Figure~\ref{fig:synth_arms_regret} further motivates per-arm adaptation using a synthetic instance where half the arms have low variance ($\sigma \in [0.2, 0.4]$) and the other half have high variance ($\sigma \in [2.0, 4.0]$). In this heterogeneous setting, neither UCB1 nor UCB-V2 dominates: UCB1 is penalized on low-variance arms, while UCB-V2 over-explores high-variance arms. UCB-minV2 achieves the lowest regret by adapting its strategy per arm, outperforming all other tested algorithms by the end of all iterations. Figure~\ref{fig:synth_arms_selection} shows the resulting selection: the
variance-aware bonus is chosen on $80$-$100\%$ of the pulls of the low-variance arms,
against $12$-$23\%$ on the high-variance ones, letting UCB-minV2 reach the
lowest final regret by following the variance profile of each arm.

\begin{figure}[H]
  \centering
  \setlength{\abovecaptionskip}{3pt}
  \captionsetup[subfigure]{skip=1pt}
  \includegraphics[width=\textwidth]{pics/replot/legend_all.pdf}\\[2pt]
  \begin{subfigure}[t]{0.49\textwidth}
    \centering
    \includegraphics[width=\linewidth]{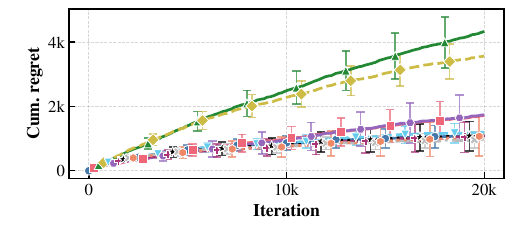}
    \caption{$w=(0.2,0.8)$.}\label{fig:synth_weights_big}
  \end{subfigure}\hfill
  \begin{subfigure}[t]{0.49\textwidth}
    \centering
    \includegraphics[width=\linewidth]{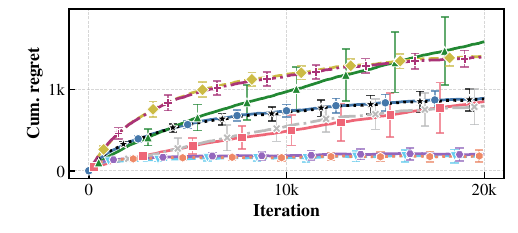}
    \caption{$w=(0.8,0.2)$.}\label{fig:synth_weights_short}
  \end{subfigure}
  \caption{Impact of scalarization weights on cumulative regret. Dim~1 has $\sigma \in [0.1, 0.5]$ (low variance), Dim~2 has $\sigma \in [2.0, 4.0]$ (high variance).}
  \label{fig:synthetic_weights}
\end{figure}

Figure~\ref{fig:synthetic_weights} shows how the choice of scalarization weights affects the variance regime and thus the relative performance of the algorithms. Here, one objective dimension has low variance ($\sigma \in [0.1, 0.5]$) and the other has high variance ($\sigma \in [2.0, 4.0]$). With $w=(0.2,0.8)$ (Fig.~\ref{fig:synth_weights_big}), the weight vector emphasizes the high-variance dimension, resulting in large scalarized variances. In this regime, UCB1 performs better while UCB-V2 over-explores due to large variance-dependent bonuses. Conversely, with $w=(0.8,0.2)$ (Fig.~\ref{fig:synth_weights_short}), the weight vector emphasizes the low-variance dimension, reducing the scalarized variance and shifting the regime toward one where variance-aware methods perform better. In both cases, UCB-minV2 tracks the better-performing base algorithm, adapting to the effective variance regime induced by the weights.

\begin{figure}[!htb]
  \centering
  \setlength{\abovecaptionskip}{3pt}
  \setlength{\belowcaptionskip}{0pt}
  \includegraphics[width=\textwidth]{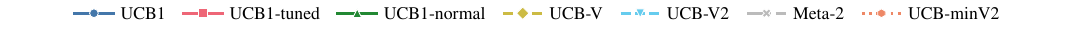}\\[2pt]
  \begin{subfigure}[t]{0.325\textwidth}
    \centering
    \includegraphics[width=\linewidth]{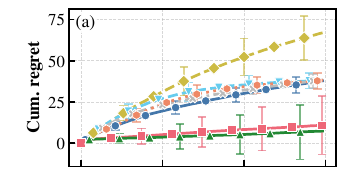}%
    \phantomsubcaption\label{fig:edu-regret-A}
  \end{subfigure}\hfill
  \begin{subfigure}[t]{0.325\textwidth}
    \centering
    \includegraphics[width=\linewidth]{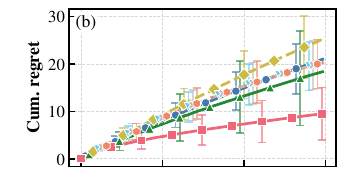}%
    \phantomsubcaption\label{fig:edu-regret-AP}
  \end{subfigure}\hfill
  \begin{subfigure}[t]{0.325\textwidth}
    \centering
    \includegraphics[width=\linewidth]{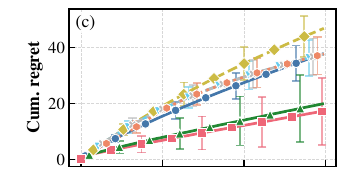}%
    \phantomsubcaption\label{fig:educ_regret_acg}
  \end{subfigure}\\[3pt]
  \begin{subfigure}[t]{0.325\textwidth}
    \centering
    \includegraphics[width=\linewidth]{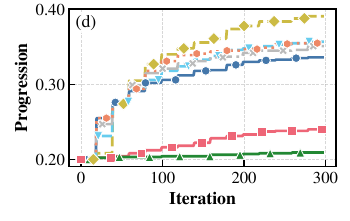}%
    \phantomsubcaption\label{fig:edu-prog-A}
  \end{subfigure}\hfill
  \begin{subfigure}[t]{0.325\textwidth}
    \centering
    \includegraphics[width=\linewidth]{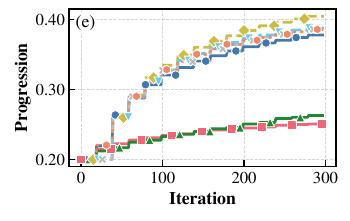}%
    \phantomsubcaption\label{fig:edu-prog-AP}
  \end{subfigure}\hfill
  \begin{subfigure}[t]{0.325\textwidth}
    \centering
    \includegraphics[width=\linewidth]{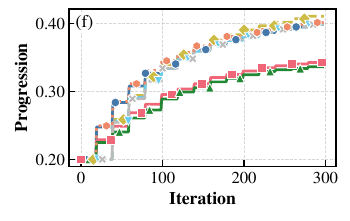}%
    \phantomsubcaption\label{fig:educ_prog_acg}
  \end{subfigure}
  \caption{Educational data: regret (top row) and progression skill (bottom row) for $T=300$. Columns correspond to different objective sets (A, A+P, A+C+G).}
  \label{fig:education_short}
\end{figure}

\paragraph*{Educational dataset (short horizon)} For this dataset, $b=1$ matches the true support since all rewards lie in $[0,1]$. Figure~\ref{fig:education_short} reports both cumulative regret and the progression skill for $T=300$ (short horizon, realistic for exercise recommendation). Comparing columns (from left to right), progression skill increases, confirming the benefit of multi-objective formulations for this task. In terms of regret, UCB1 tends to perform better in the very early stages, while the variance-aware UCB-V2 becomes more effective later as more observations are collected and variance estimates stabilize. UCB-minV2 shows a similar transition: it behaves like UCB1 in the initial iterations (e.g.\ the first 50 in Fig.~\ref{fig:educ_regret_acg}) before shifting toward UCB-V2.

\begin{figure}[!htb]
  \centering
  \setlength{\abovecaptionskip}{3pt}
  \includegraphics[width=\textwidth]{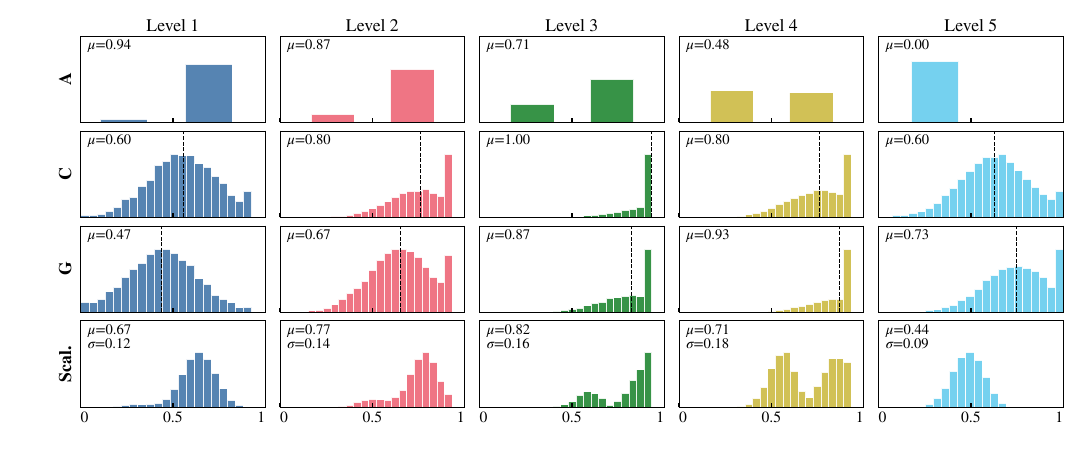}
  \caption{Reward distributions per arm for a given user (level~3) in the A+C+G setting with equal weights. Rows: individual objective dimensions and scalarized reward. Columns: difficulty levels (arms).}
  \label{fig:educ_distributions}
\end{figure}

Figure~\ref{fig:educ_distributions} shows the reward distributions of one user across all arms in the A+C+G setting with equal weights, one row per objective dimension and the scalarized reward on the bottom row. The dimensions have very different shapes, Aptitude being binary and Competence and Gap continuous with arm- and user-dependent moments. After scalarization the resulting distributions differ across arms in both mean and variance, and their shape is arm-specific and hard to anticipate: some are close to Gaussian, others bi-modal. This motivates an algorithm that adapts its exploration per arm from observed data rather than from assumptions on the underlying distributions.

\paragraph*{Group recommendation - MovieLens and Steam} These experiments use the two-objective formulation of~\cite{amer2009group} (relevance and pairwise disagreement) and show that the proposed meta-algorithm adaptively follows the best underlying base strategy.

\begin{figure}[!htb]
  \centering
  \setlength{\abovecaptionskip}{3pt}
  \setlength{\belowcaptionskip}{0pt}
  \includegraphics[width=\textwidth]{pics/replot/legend_some.pdf}\\[2pt]
  \begin{subfigure}[t]{0.325\textwidth}
    \centering
    \includegraphics[width=\linewidth]{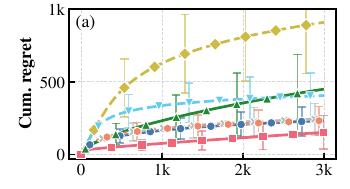}%
    \phantomsubcaption\label{fig:regret_movielens}
  \end{subfigure}\hfill
  \begin{subfigure}[t]{0.325\textwidth}
    \centering
    \includegraphics[width=\linewidth]{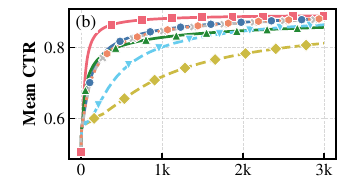}%
    \phantomsubcaption\label{fig:ctr_movielens}
  \end{subfigure}\hfill
  \begin{subfigure}[t]{0.325\textwidth}
    \centering
    \includegraphics[width=\linewidth]{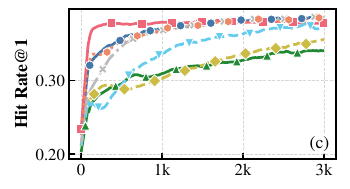}%
    \phantomsubcaption\label{fig:hit_movielens}
  \end{subfigure}\\[3pt]
  \begin{subfigure}[t]{0.325\textwidth}
    \centering
    \includegraphics[width=\linewidth]{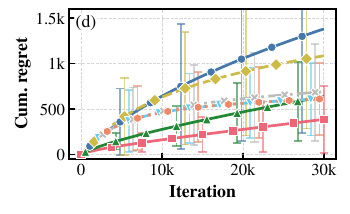}%
    \phantomsubcaption\label{fig:regret_steam}
  \end{subfigure}\hfill
  \begin{subfigure}[t]{0.325\textwidth}
    \centering
    \includegraphics[width=\linewidth]{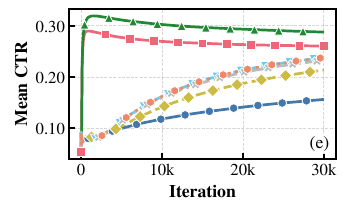}%
    \phantomsubcaption\label{fig:ctr_steam}
  \end{subfigure}\hfill
  \begin{subfigure}[t]{0.325\textwidth}
    \centering
    \includegraphics[width=\linewidth]{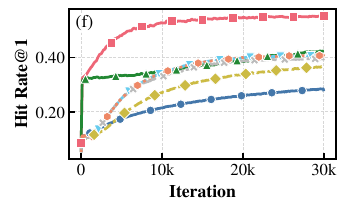}%
    \phantomsubcaption\label{fig:hit_steam}
  \end{subfigure}
  \caption{MovieLens (top row) and Steam (bottom row): Relevance and Pairwise Disagreement. Error bars are clipped in (a) and (d).}
  \label{fig:group_relxapd}
\end{figure}

On MovieLens (Figs.~\ref{fig:regret_movielens}-\ref{fig:hit_movielens}), rewards lie in $[0,5]$ (not normalized), so $b=1$ underestimates the true support. UCB1's bonus is then moderate and provides sufficient exploration, while UCB-V2 over-explores as its bonus scales with the empirical variances, which are large ($>1$). UCB-minV2 identifies this regime and aligns with UCB1 throughout the horizon, also matching it on CTR (Fig.~\ref{fig:ctr_movielens}) and Hit Rate@1 (Fig.~\ref{fig:hit_movielens}).

On Steam (Figs.~\ref{fig:regret_steam}-\ref{fig:hit_steam}), rewards are in $[0,1]$ and per-arm variances are small, a regime where variance-aware exploration is optimal: among UCB1, UCB-V, and UCB-V2, UCB-V2 achieves the lowest regret (Fig.~\ref{fig:regret_steam}) and UCB-minV2 closely tracks it, consistently across all metrics (Figures~\ref{fig:ctr_steam}-\ref{fig:hit_steam}); UCB1-Tuned reaches a lower regret on this dataset. This contrast with MovieLens illustrates the dataset-dependent nature of the optimal exploration strategy and the robustness of UCB-min.

\begin{table}[!t]
  \centering
  \caption{Final cumulative regret at the horizon $T$ of each experiment (mean over the 50 runs), computed from the plotted curves of Figs.~\ref{fig:synthetic}, \ref{fig:education_short}, and~\ref{fig:group_relxapd}; lowest value per column in bold. UCB1-Tuned, Meta-1, and Meta-2 are not reported, as they have no regret guarantee. Synthetic: low-variance, high-variance, and heterogeneous regimes (Figs.~\ref{fig:synth_small}-\ref{fig:synth_arms_regret}), with $T=20$k, $20$k, and $40$k. Education: objective sets A, A+P, and A+C+G. ``-'': not shown in the figures of this dataset.}
  \label{tab:final_regret}
  \setlength{\tabcolsep}{5pt}
  \begin{tabular}{lrrrrrrrr}
    \toprule
    & \multicolumn{3}{c}{Synthetic} & \multicolumn{3}{c}{Education ($T=300$)} & MovieLens & Steam \\
    \cmidrule(lr){2-4}\cmidrule(lr){5-7}
    Algorithm & Low & High & Hetero. & A & A+P & A+C+G & $T=3$k & $T=30$k \\
    \midrule
    UCB1        & 1005.5 & 1064 & 1373 & 38.0 & 21.1 & 37.8 & 234 & 1381 \\
    UCB1-Normal & 271.7 & 4124 & 6490 & \textbf{7.4} & \textbf{18.4} & \textbf{19.9} & 450 & 638 \\
    UCB-V       & 776.9 & 3248 & 3729 & 67.3 & 25.3 & 46.9 & 908 & 1086 \\
    UCB-V2      & 94.5 & 2323 & 851 & 38.4 & 20.4 & 37.7 & 406 & 613 \\
    EUCBV       & 172.9 & 1621 & 965 & - & - & - & - & - \\
    UCB-minV1   & 742.0 & 969 & 1267 & - & - & - & - & - \\
    UCB-minV2   & \textbf{92.6} & \textbf{958} & \textbf{799} & 38.2 & 20.4 & 37.7 & \textbf{233} & \textbf{612} \\
    \bottomrule
  \end{tabular}
\end{table}

\begin{figure}[!htb]
  \centering
  \setlength{\abovecaptionskip}{3pt}
  \setlength{\belowcaptionskip}{0pt}
  \includegraphics[width=0.83\textwidth]{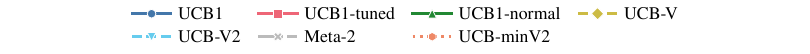}\\[2pt]
  \begin{subfigure}[t]{0.4\textwidth}
    \centering
    \includegraphics[width=\linewidth]{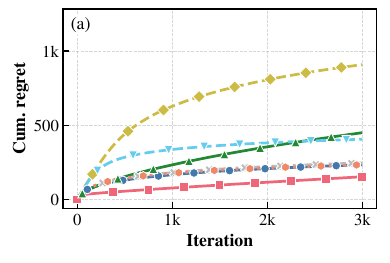}%
    \phantomsubcaption\label{fig:norm_raw}
  \end{subfigure}\hspace{0.02\textwidth}
  \begin{subfigure}[t]{0.4\textwidth}
    \centering
    \includegraphics[width=\linewidth]{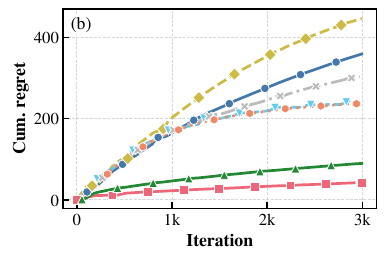}%
    \phantomsubcaption\label{fig:ml_normalized}
  \end{subfigure}
  \caption{Effect of normalization (a: raw rewards in $[0,5]$, b: normalized to $[0,1]$) on MovieLens cumulative regret.}
  \label{fig:normalization}
\end{figure}

\paragraph*{Effect of Reward Normalization on MovieLens}
Figures~\ref{fig:norm_raw}-\ref{fig:ml_normalized} compare UCB-type algorithms on MovieLens under raw rewards in $[0,5]$ and rewards normalized to $[0,1]$ by dividing by the maximum rating, with $b=1$ in both cases.

The two settings induce different variance regimes. With raw rewards, empirical variances are large ($>1$), placing the problem in a high-variance regime where UCB1 outperforms UCB-V2, which over-explores. After normalization, variances are reduced by a factor of 25 and UCB-V2 becomes competitive. In both cases, UCB-minV2 identifies the prevailing regime and tracks the better base algorithm throughout the horizon, confirming the robustness of the per-arm minimum rule.

Normalization does not improve the best achievable cumulative regret either. It is thus not a straightforward remedy to variance heterogeneity and, critically, requires knowing the reward upper bound in advance, which is unavailable online.

\begin{figure}[!htb]
  \centering
  \setlength{\abovecaptionskip}{3pt}
  \setlength{\belowcaptionskip}{0pt}
  \includegraphics[width=0.83\textwidth]{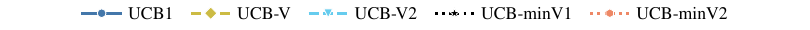}\\[2pt]
  \begin{subfigure}[t]{0.4\textwidth}
    \centering
    \includegraphics[width=\linewidth]{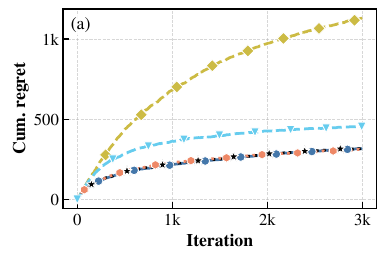}%
    \phantomsubcaption\label{fig:b1}
  \end{subfigure}\hspace{0.02\textwidth}
  \begin{subfigure}[t]{0.4\textwidth}
    \centering
    \includegraphics[width=\linewidth]{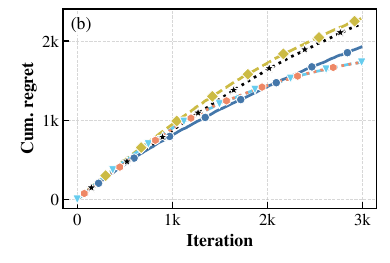}%
    \phantomsubcaption\label{fig:b5}
  \end{subfigure}
  \caption{Effect of the support bound $b$ (a: $b=1$, b: $b=5$, one group) on MovieLens cumulative regret.}
  \label{fig:support_bound}
\end{figure}

\paragraph*{Upper Bound Reward Effect}
Figures~\ref{fig:b1}-\ref{fig:b5} compare the effect of the support bound $b$ on MovieLens regret: $b=1$ (used throughout) vs.\ $b=5$ (the true support $[0,5]$). With $b=1$, UCB1 and UCB-minV2 reach substantially lower regret than UCB-V2, whose variance-dependent bonus already over-explores. With $b=5$, all bonuses grow proportionally and every method reaches near-linear regret, confirming that a large $b$, even well-specified, penalizes all UCB-type algorithms alike, variance-aware or not, and motivating $b=1$ as a default. The minimum rule is especially useful here: even when $b$ underestimates the support, it picks the tighter per-arm bonus.

\section{Conclusion}
This paper studied exploration in bandits whose arms have different variance regimes. We showed that the scalarization step in multi-objective optimization can lead to different variance levels across arms, making the choice between variance-independent and variance-aware exploration both arm- and instance-dependent. Neither UCB1 nor UCB-V2 uniformly dominates the other: their relative performance depends on the variance regime, the covariance structure of the objectives, and the reward support bound. We proposed \textsc{UCB-min}, a per-arm adaptive meta-algorithm that selects, for each arm and at each round, the tighter of the two exploration bonuses. It requires no prior knowledge of the variance regime, no reward normalization, and is empirically robust to an underestimated reward bound. Experiments on synthetic data, an educational tutoring task, and two group recommendation benchmarks confirmed that it consistently tracks the best base algorithm across diverse variance regimes and datasets.

Future work includes characterizing when minimum-of-bounds meta-algorithms achieve best-of-both-worlds regret, in particular whether the guarantee holds at the standard $c=1.2$ of~\cite{audibert2007tuning} and whether UCB-minV2's strong empirical behavior admits a matching bound. Other directions include experiments with other weight vectors and more objectives on the real-world datasets.

\section*{Acknowledgment}

This work was partly supported by DataGEMS, funded by the European Union's Horizon Europe Research and Innovation Programme, under grant agreement number 101188416.

Generative AI tools were used for language editing and for suggesting related work, coding, and initial proof ideas; all technical content and results are the authors' own.

\bibliographystyle{IEEEtran}
\bibliography{biblio}

\appendices
\setlength{\abovedisplayskip}{6pt}
\setlength{\belowdisplayskip}{6pt}
\setlength{\abovedisplayshortskip}{3pt}
\setlength{\belowdisplayshortskip}{3pt}

\section{Proof of Theorem~\ref{thm:main}}\label{apx:proof}

\begin{proof}
Consider a UCB bandit $\pi$ with indexes $B_{a,s,t}$, where $B_{a,s,t}$ is the index of arm $a$ at step $t$ when $a$ has been pulled $s$ times (here $s$ counts pulls, unrelated to the superscript $(s)$ of scalarized quantities), with $a^\star$ an optimal arm and $\mu^\star$ its expected reward. The approach of~\cite{auer2002finite,audibert2009exploration} to bound the expected regret $R(T)$ of UCB1 (resp. UCB-V) is to decompose it as $R(T)=\sum_a \Delta_a\mathbb{E}(N_a(T))$. They bound $\mathbb{E}(N_a(T))$ for suboptimal arms by noting that, under the UCB rule, arm $a$ is pulled at step $t$ only if its index is at least that of $a^\star$, which requires one of two ``bad'' events: (i) the index of $a$ is larger than $\mu^\star$, or (ii) the index of $a^\star$ is smaller than $\mu^\star$. Fixing some value $u$ to be tuned later and bounding the probability of the two events for each step $t$ after $a$ has been pulled $u$ times, any UCB bandit $\pi$ must satisfy:

\begin{multline}\label{eq:decompose-nbpulls-threshold}
\mathbb{E}(N_a(T)) \leq u + \sum_{t=u+1}^T\mathbb{P}\bigl(\exists u\leq s\leq t-1  : B_{a,s,t} > \mu^\star\bigr) \\
+ \sum_{t=u+1}^T\mathbb{P}\bigl(\exists 1\leq s\leq t-1  :  B_{a^\star,s,t} \leq \mu^\star\bigr)
\end{multline}

This equation appears in slightly different forms, each upper bounding every term of Eq.~\eqref{eq:decompose-nbpulls-threshold}, in~\cite[Theorem 2]{audibert2009exploration} and~\cite[Proof of Theorem~1]{auer2002finite}; both papers bound the expected regret of UCB1 and UCB-V by bounding each term of the sum.

We will denote by $u_a^{(1)}$, $B_{a,s,t}^{(1)}$, $U_a^{(1)}(T)$ the threshold value, index, and upper bound on $\mathbb{E}(N_a(T))$ obtained in \cite{auer2002finite} for UCB1, and will use similarly $u_a^{(V)}$, $B_{a,s,t}^{(V)}$ and $U_a^{(V)}(T)$ for UCB-V, while the version without superscript will be used for UCB-minV1.
So it suffices to prove that 
$\mathbb{E}(N_a(T)) \leq U_a^{(1)}(T) + U_a^{(V)}(T)$ to establish our regret bound for UCB-minV1 with $c = 1.2$.

\begin{align*}
u_a^{(1)} &= \left\lceil 8b^2\log T/\Delta_a^2 \right\rceil,\\
u_a^{(V)} &= \left\lceil 8 \left(\tfrac{\sigma_a^2}{\Delta_a^2} + \tfrac{2b}{\Delta_a}\right) c \log T \right\rceil .
\end{align*}

Furthermore by monotonicity of the left expression w.r.t. $B$:
\begin{multline}
\label{eq:metas-suboptimal-overshoot-bounded-by-min}
\mathbb{P}\bigl(\exists u\leq s\leq t-1  : B_{a,s,t} > \mu^\star\bigr) \leq \\
\min\Bigl(
\mathbb{P}\bigl(\exists u\leq s\leq t-1  : B_{a,s,t}^{(1)} > \mu^\star\bigr), \\
\mathbb{P}\bigl(\exists u\leq s\leq t-1  : B_{a,s,t}^{(V)} > \mu^\star\bigr)  
\Bigr)
\end{multline}
 
and by union bound
\begin{multline}\label{eq:metas-proba-optimals-confidbound-fails-bounded-by-sum}
\mathbb{P}\bigl(\exists s\leq t-1  :  B_{a^\star,s,t} \leq \mu^\star\bigr)
 \leq \\
 \mathbb{P}\bigl(\exists s\leq t-1  :  B_{a^\star,s,t}^{(1)} \leq \mu^\star\bigr)\\
 + \mathbb{P}\bigl(\exists s\leq t-1  :  B_{a^\star,s,t}^{(V)} \leq \mu^\star\bigr)
\end{multline}
Since~\cite{auer2002finite} and~\cite{audibert2009exploration} obtain their bound by bounding all the probabilities appearing in the sums for their respective choice of $u$, combining Equations~\eqref{eq:decompose-nbpulls-threshold},
\eqref{eq:metas-suboptimal-overshoot-bounded-by-min} and \eqref{eq:metas-proba-optimals-confidbound-fails-bounded-by-sum} with $u = \max\bigl(u_a^{(1)} , u_a^{(V)}\bigr)$ yields a sum whose terms all appear in the corresponding sums for UCB1 or UCB-V. The expected number of pulls of suboptimal arms in UCB-minV1, hence its expected regret, is therefore at most the sum of those for UCB1 and UCB-V, which gives the result for $c=1.2$ in the Theorem.

For the tighter result, we prove that for $c\geq4$ UCB-minV1 does not pull arm $a$ more than a few times above the best of the two bounds above, namely $\mathbb{E}(N_a(T)) \leq \min\bigl(U_a^{(1)}(T), U_a^{(V)}(T)\bigr) + O(1)$. We first bound by a constant the last sum in Eq.~\eqref{eq:decompose-nbpulls-threshold}, then bound the first two terms.

Recall from~\cite{auer2002finite} that by Chernoff-Hoeffding, $\mathbb{P}(B_{a^\star,s,t}^{(1)} \leq \mu^\star) \leq t^{-4}$ for all $s$ and $t$.
When $c\geq 4$, the empirical Bernstein inequality \cite[Eq. (7) of Theorem~1]{audibert2009exploration} provides a similar concentration bound for UCB-V, so we can assume $\mathbb{P}(B_{a^\star,s,t}^{(V)} \leq \mu^\star) \leq 3t^{-4}$.
For any choice of $u$, we next show that the last term of Eq.~\eqref{eq:decompose-nbpulls-threshold} is bounded by a constant. For all $t$ and $s$, we have by union bound:
\begin{equation*}
\mathbb{P}(B_{a^\star,s,t} \leq \mu^\star)\leq \mathbb{P}(B_{a^\star,s,t}^{(1)} \leq \mu^\star)+\mathbb{P}(B_{a^\star,s,t}^{(V)} \leq \mu^\star)\leq 4t^{-4}.
\end{equation*}
Consequently and decomposing by union bound: 
\begin{equation*}
\begin{split}
&\sum_{t=u+1}^T\mathbb{P}(\exists 1\leq s\leq t-1  :  B_{a^\star,s,t} \leq \mu^\star)\\
&\quad\leq \sum_{t=u+1}^T\sum_{s=1}^{t-1} \mathbb{P}(B_{a^\star,s,t} \leq \mu^\star)
\leq \sum_{t=u+1}^T\sum_{s=1}^{t-1} 4t^{-4}\end{split} 
\end{equation*}
This proves that 
\begin{equation}\label{eq:contrib-ucb-optimalarm-fails-is-constant}
    \sum_{t=u+1}^T\mathbb{P}(\exists 1\leq s\leq t-1  :  B_{a^\star,s,t} \leq \mu^\star) \in O(1)
\end{equation}

To fix a choice for $u$, we compare the two bounds:
if $U_a^{(1)} \leq U_a^{(V)}$, we set 
$u = u_a^{(1)}$. Otherwise, we set $u = u_a^{(V)}$.
As we are only summing terms that appear in the smallest of the two bounds, possibly replacing some of these terms by smaller terms from the other bound (according to Eq.~\eqref{eq:metas-suboptimal-overshoot-bounded-by-min}),
\begin{equation*}
    u + \sum_{t=u+1}^T \mathbb{P}\bigl(\exists u\leq s\leq t-1  : B_{a,s,t} > \mu^\star\bigr) \leq \min\bigl(U_a^{(1)}, U_a^{(V)}\bigr)
\end{equation*}
Combining with Eq.~\eqref{eq:contrib-ucb-optimalarm-fails-is-constant}, we deduce that
\begin{equation*}
    \mathbb{E}(N_a(T)) \leq \min\bigl(U_a^{(1)}(T), U_a^{(V)}(T)\bigr)+O(1)
\end{equation*}
which concludes our proof of the tighter result in the Theorem, obtained for $c\geq 4$.
\end{proof}

\section{Metrics description}\label{apx:metrics}

\paragraph{Bandit performance} 
\begin{itemize}
    \item \textbf{Cumulative Regret}, as defined in Eq.~\eqref{eq:regret}, measures decision suboptimality (\emph{lower is better}).
\end{itemize}

\paragraph{Ranking and interaction} At round $t$, let $\mathcal{L}_t=(a_1,\dots,a_k)$ be the recommendation list (last $k$ recommendations) and $\mathcal{S}\subseteq\mathcal{A}$ the truly relevant arms (ground truth). For $k\in\{1,3,5\}$ we define \textbf{Precision@k}$=|\{a_1,\dots,a_k\}\cap\mathcal{S}|/k$ and \textbf{Hit Rate@k}$=\mathbf{1}(|\{a_1,\dots,a_k\}\cap\mathcal{S}|>0)$; the paper reports Hit Rate@1, the remaining values being available online.\textsuperscript{\ref{fn:github}} These assess whether relevant items appear among the last $k$ recommendations; we take $\mathcal{S}$ as the top-$M$ arms by true scalar mean $\mu_a^{(s)}$, with $M=5$.

\paragraph{Engagement and domain}

\textbf{CTR}$=\frac{1}{T}\sum_{t=1}^T \mathbf{1}\{\mathrm{rel}_t>0\}$, where $\mathrm{rel}_t$ is the observed relevance of the recommended item at $t$ (\emph{higher is better}); \textbf{Skill progression (NCC)}: number of difficulty levels mastered, a level requiring $N$ consecutive correct responses (\emph{higher is better}).

\end{document}